\documentclass[10pt,twocolumn,letterpaper]{article}

\usepackage[pagenumbers]{cvpr} 
\usepackage{multirow}
\definecolor{cvprblue}{rgb}{0.21,0.49,0.74}
\usepackage[pagebackref,breaklinks,colorlinks,allcolors=cvprblue]{hyperref}

\def\paperID{*****} 
\def\confName{}
\def\confYear{2025}

\title{Gaussian Pixel Codec Avatars: A Hybrid Representation for Efficient Rendering}

\author{
Divam Gupta$^{3}$\thanks{Work done at Meta.}, Anuj Pahuja$^{1}$, Nemanja Bartolovic$^{1}$, Tomas Simon$^{1}$, Forrest Iandola$^{2}$, Giljoo Nam$^{1}$ \\
$^{1}$Meta Codec Avatars Lab \\
$^{2}$Meta Reality Labs \\
$^{3}$Stellon Labs
}

\begin{document}

\twocolumn[{%
\maketitle

\renewcommand\twocolumn[1][]{#1}%
\maketitle
\begin{center}
\centering
\vspace{-20pt}
\captionsetup{type=figure}
\includegraphics[width=\linewidth]{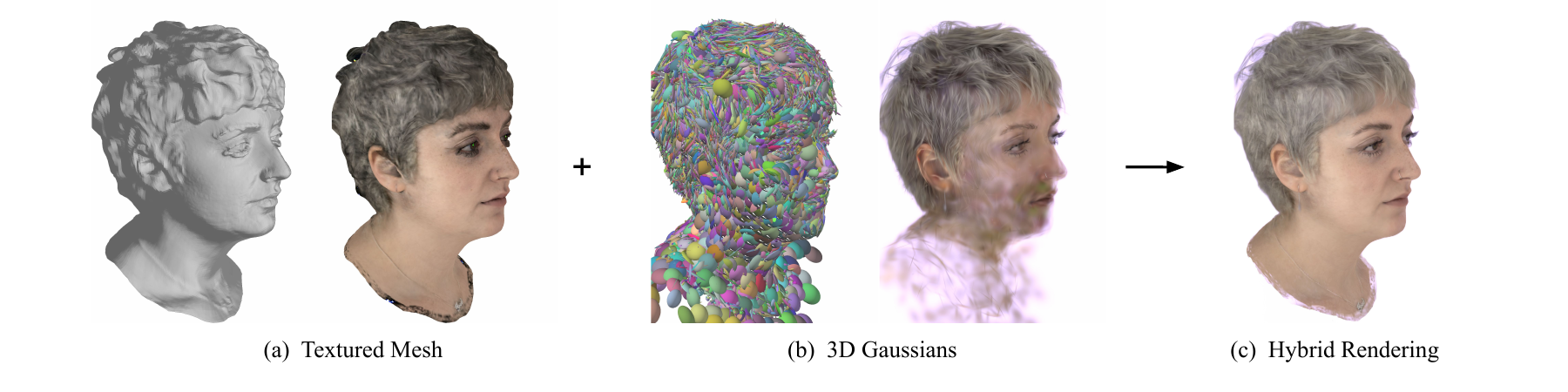} 
\vspace{-22pt}
\caption{\textbf{ Gaussian Pixel Codec Avatars.} We present Gaussian Pixel Codec Avatars (GPiCA), photorealistic head avatars that combine (a) a textured triangle mesh, with (b) 3D Gaussians, to produce a (c) hybrid rendering representation that retains high realism while being efficient to render even on low-end devices}
\label{fig:teaser}
\end{center}
}]

\begingroup
\renewcommand{\thefootnote}{\fnsymbol{footnote}}
\footnotetext[1]{Work done at Meta.}
\endgroup

\begin{abstract}
We present Gaussian Pixel Codec Avatars (GPiCA), photorealistic head avatars that can be generated from multi-view images and efficiently rendered on mobile devices. GPiCA utilizes a unique hybrid representation that combines a triangle mesh and anisotropic 3D Gaussians. This combination maximizes memory and rendering efficiency while maintaining a photorealistic appearance. The triangle mesh is highly efficient in representing surface areas like facial skin, while the 3D Gaussians effectively handle non-surface areas such as hair and beard. To this end, we develop a unified differentiable rendering pipeline that treats the mesh as a semi-transparent layer within the volumetric rendering paradigm of 3D Gaussian Splatting. We train neural networks to decode a facial expression code into three components: a 3D face mesh, an RGBA texture, and a set of 3D Gaussians. These components are rendered simultaneously in a unified rendering engine. The networks are trained using multi-view image supervision. Our results demonstrate that GPiCA achieves the realism of purely Gaussian-based avatars while matching the rendering performance of mesh-based avatars.
\end{abstract}


\section{Introduction}

\begin{figure*}
\centering
\includegraphics[width=\textwidth]{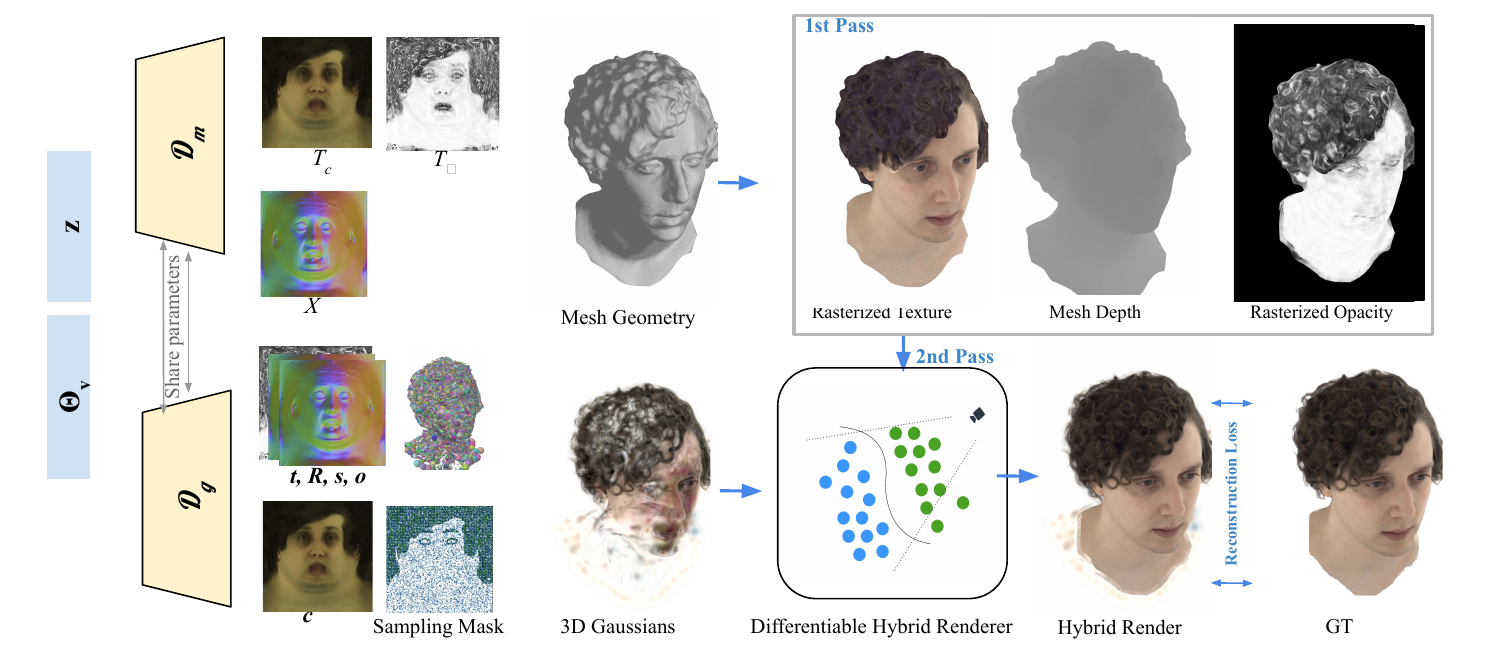}
\caption{\textbf{Overview.} Given an expression latents code and view direction, our model outputs 3D Gaussians and a mesh in the UV texture space. The Gaussian positions are predicted relative to the mesh. A sampling mask is used to select 3D Gaussians for key regions like hair, beard etc. In the first pass, the mesh is rasterized to produce RGB texture, opacity and depth. In the second pass the 3D Gaussians, rasterized texture, depth and opacity are sent to our differentiable hybrid renderer. The renderer first accumulates colors of Gaussians in front of the mesh, then accumulate the mesh component and finally accumulates Gaussians behind the mesh. The system is trained jointly. }
\label{fig:your_label}
\end{figure*}

Modeling, animating, and rendering human faces have been an important topic of interest in computer graphics for many years~\cite{parke1972computer,debevec2000acquiring}. Recently, there have been remarkable advancements in creating animatable head avatars~\cite{wang2022morf,sarkar2023litnerf,kabadayi24ganavatar,saito2024rgca}. These achievements can be attributed to three key components: large-scale face datasets, learning-based approaches, and efficient representations. Firstly, multi-view video capture systems~\cite{wuu2022multiface,kirschstein2023nersemble} have enabled the collection of extensive datasets of faces in motion, allowing for fully data-driven methods that can model subtle expressions. Secondly, deep generative models such as GAN~\cite{goodfellow2020generative} and VAE~\cite{kingma2013auto} have made it possible to learn compact latent spaces of geometry and appearance of faces in motion. Lastly, advances in scene representations, such as Neural Radiance Field (NeRF)~\cite{mildenhall2021nerf} and 3D Gaussian Splatting (3DGS)~\cite{kerbl20233d}, have provided the means for photorealistic modeling and rendering of avatars. This paper focuses on the representation aspect of head avatars and proposes a new hybrid representation combining triangle meshes and 3D Gaussians for efficient rendering.

Why does representation matter?  First, for head avatars, where accurately capturing shape and appearance is essential, the type of representation being used can limit expressiveness. Triangle meshes are a popular choice in computer graphics due to their simplicity and efficient rendering. However, it is challenging to accurately reconstruct images of hair into a mesh. Second, representations directly impact rendering performance. This is crucial for real-time AR/VR applications like telepresence, where latency has to be minimal while being compute limited. Therefore, we aim to find a representation that is efficient enough to run on mobile devices while maintaining high fidelity in reproducing complex human face geometry and appearance.

We present Gaussian Pixel Codec Avatars (GPiCA), photorealistic head avatars learned from multi-view images. The gist of GPiCA is a hybrid representation of a triangle mesh and anisotropic 3D Gaussians that maximizes the memory and rendering efficiency while achieving photorealistic appearance, especially around hair.
The mesh can efficiently represent surface areas like facial skin whereas 3D Gaussians deal with non-surface areas like hair and beard.
To this end, we develop a unified differentiable rendering pipeline that views the mesh as a semi-transparent layer within the volumetric rendering paradigm of 3DGS.
With that, we train neural networks that decode a facial expression code into a 3D face mesh, a RGBA texture, and a set of 3D Gaussians.
All the three components are rendered in a unified rendering engine at once. 
The networks are trained via multi-view image supervision. 
We show that GPiCA is as efficient as mesh-based avatars in terms of rendering performance and is as realistic as purely Gaussian-based avatars. 



\section{Related Work}

\subsection{Representations for Face Modeling}
\paragraph{Mesh}
The triangle mesh is one of the most widely used representations for surfaces in computer graphics, and modern GPUs can rasterize billions of triangles per second. In addition to its exceptional computational efficiency, meshes also provide a strong benefit to face modeling; the shared topology enables to build dense correspondences across different identities~\cite{cao2013facewarehouse,gerig2018morphable}. This has been shown in the success of various 3D morphable face models (3DMM), most of which are based on triangle meshes~\cite{egger20203d}. Such facial 3D template models~\cite{li2017learning} have been used to create photorealistic mesh-based head avatars using various datasets such as selfies~\cite{grassal2022neural}, in-the-wild videos~\cite{khakhulin2022realistic}, or multi-view face captures~\cite{lombardi2018deep,ma2021pixel}. While being efficient and robust, meshes are reported to have limited capabilities to faithfully reproduce the complex appearance of hair. 

\paragraph{Volume}
Volumes are good alternatives to meshes for non-surface areas like hair thanks to its flexibility to learn geometry and appearance from images. It has been shown that volumes are effective representations for learning 3D faces from unstructured large-scale portrait images~\cite{chan2022efficient,sun2023next3d} or creating animatable head avatars from structured multi-view face captures~\cite{lombardi2019neural}. Volumetric representations are often combined with mesh-based 3DMMs for the sake of face registration~\cite{zielonka2023instant}, tracking~\cite{zheng2022imavatar}, and animation~\cite{lombardi2021mixture}. While volumes usually achieve noticeable visual improvements in the areas like hair, beard, and eyebrow, its high computational cost can be a problem in realtime applications with a budget constrained environment such as mobile AR/VR devices. 


\paragraph{3D Gaussian Splatting}
3D Gaussian Splatting (3DGS)~\cite{kerbl20233d} employs anisotropic 3D Gaussians as a scene representation and achieves realtime rendering performance via software rasterization. A number of works have demonstrated the effectiveness of 3DGS as a representation of head avatars~\cite{qian2023gaussianavatars,xu2023gaussianheadavatar,zhao2024psavatar,SplattingAvatar:CVPR2024,xiang2024flashavatar,saito2024rgca}. Similar to volumetric representations, binding 3D Gaussians with a 3DMM is a common practice for registering, tracking, and animating faces~\cite{qian2023gaussianavatars,xu2023gaussianheadavatar,zhao2024psavatar,saito2024rgca}. Despite being an efficient representation for realtime applications, 3DGS still has a trade-off between quality and memory/speed; millions of 3D Gaussians are typically required to achieve high fidelity photorealism~\cite{fan2023lightgaussian,hamdi2024ges,lee2023compact,morgenstern2023compact}. Although high-end GPUs can render millions of 3D Gaussians in realtime, achieving this within budget constrained environments, such as mobile AR/VR devices, remains a research challenge.

\paragraph{Meshes and Gaussians}
Concurrent work to ours explored a hybrid representation of a mesh and 3D Gaussians that both representations contribute to final pixel colors. Wang et al.~\cite{wang2024mega} explicitly divided head avatars into two regions, face and hair, and used mesh for face and 3DGS for hair. Unlike our work, they do not jointly train mesh and 3D Gaussians. They also treat mesh as a fully opaque layer, and they composite the mesh and 3DGS renderings via screen space alpha blending.  Xiao et al.~\cite{xiao2024bridging} presented a hybrid Gaussian-mesh rendering technique that treats the mesh as a semi-transparent layer within 3DGS rasterization framework, similar to our hybrid rendering paradigm. They focus on a dynamic scene reconstruction problem and do not focus reducing the number of Gaussians. Our primary goal is to build a head avatar that can be animated from various driving signals such as head-mount cameras in mobile VR headsets which have compute constraints. 

Svitov et al.~\cite{svitov2024haha} proposed a hybrid method for generating full body avatars from monocular videos. First, they separately generate a mesh and a Gaussian avatar.
Then in the filtering stage, they remove the 3D Gaussians which are not needed.
They always keep the mesh opaque and only update the color and opacity of the 3D Gaussians in the filtering stage.  
Because of this, the Gaussians are not able to capture details in complex areas like hair where 3D Gaussians are needed the most.
On the other hand, we jointly train a semi-transparent mesh and 3D Gaussians, because of which we are able to capture those complex areas.
Our experiments show that mesh coarsely approximates areas like hair, and having an opaque mesh hinders the 3D Gaussians.

\subsection{Meshes and Gaussians for 3D Reconstruction}
There have been efforts to relate 3D Gaussians to meshes for 3D reconstruction. A common approach is to put a hard constraint that forces 3D Gaussians to lie on the faces of a triangle mesh and update the mesh vertices during optimization processes~\cite{guedon2023sugar,chen2023neusg,waczynska2024games,lin2024direct}.
Or one can put a soft constraint that allows some thickness on a surface, so the Gaussians can better represent volumetric appearance~\cite{guedon2024frosting}. It was also shown that directly optimizing 2D oriented Gaussian disks, instead of 3D Gaussians, helps better surface reconstruction~\cite{Huang2DGS2024,Dai2024GaussianSurfels}. All these works either use meshes as a hidden representation to support the optimization of 3D Gaussians or simply extract meshes after the optimization, whereas our hybrid mesh-Gaussian rendering allows both representations to contribute to the final pixel color.

\section{Methodology}

\begin{figure}
\centering
\includegraphics [width=\linewidth]{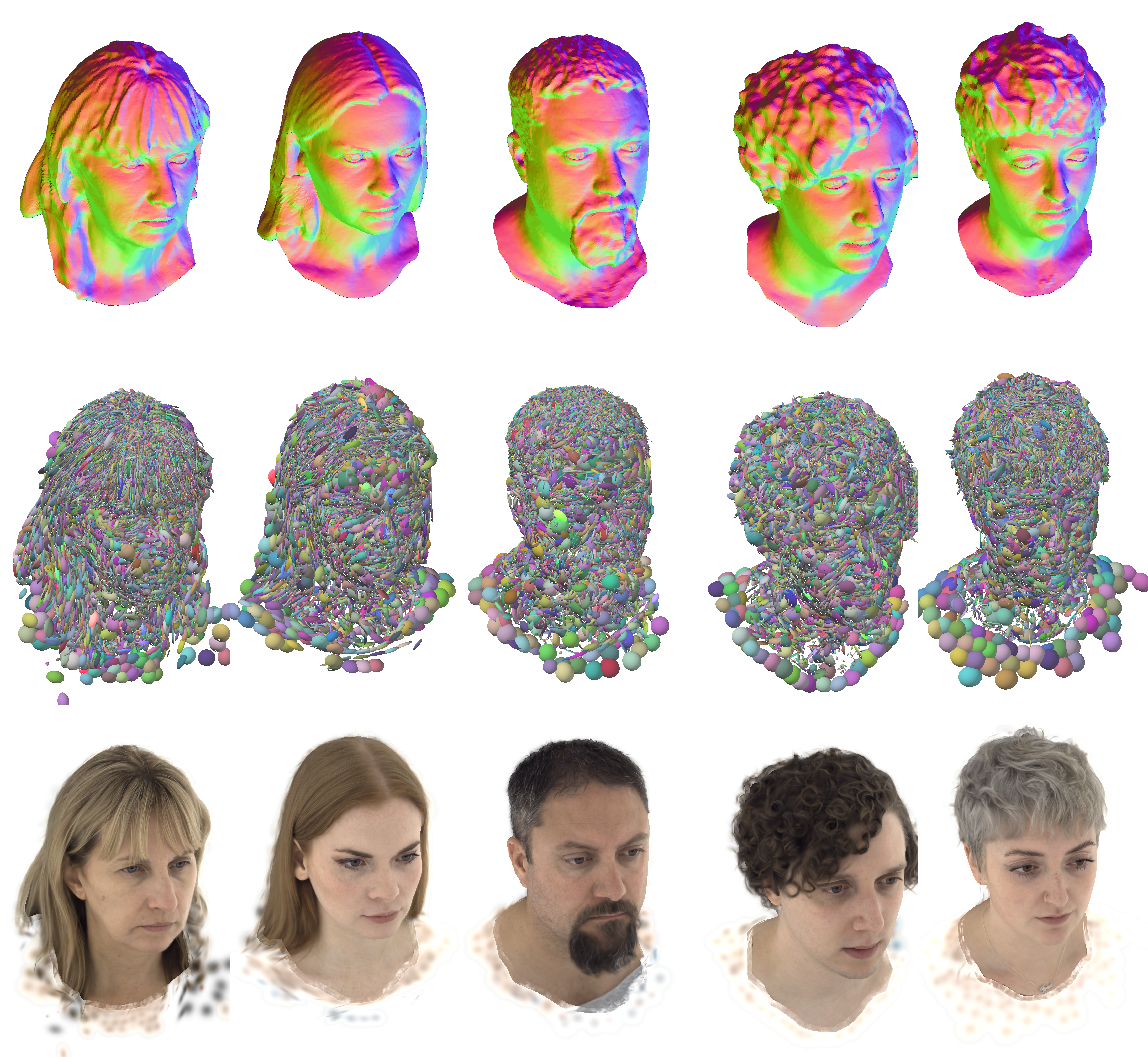}
\caption{\textbf{Learned mesh and Gaussian primitives.} \textit{Top row}: Per-vertex normals of the learned mesh. \textit{Middle row}: Learned Gaussians as 3D ellipses. \textit{Bottom row}: Final renders.}
\label{fig:normals}
\end{figure}

Our method combines a triangle mesh and anisotropic 3D Gaussians to render an avatar. The mesh captures the majority of the avatar's visual and geometrical information, while the 3D Gaussians supplement the visual data that the mesh cannot represent. These 3D Gaussians are strategically placed in areas such as head hair, facial hair, eyebrows, and eyelashes, which are beyond the representational capacity of a mesh. To render the Gaussian with a mesh, we first rasterize the mesh using differentiable rendering. Following this, we blend the visual information with the 3D Gaussians in 3D space. We refer to this as {\em hybrid mesh-Gaussian rendering}. This process is differentiable, allowing it to be learned in an end-to-end manner.

Our objective is not to reconstruct a scene, but to create an animatable avatar that can be rendered on mobile devices. To this end, we train a variational autoencoder that produces all the necessary visual and geometric information. The encoder produces a latent code $\mathbf{z}$ which serves as a conditioning input for the decoder to produce the avatar. To make our rendering efficient on mobile devices, only use 3D Gaussians where they are absolutely needed. While our method is independent of the underlying mesh model, we utilize PiCA \cite{ma2021pixel} due to its efficiency and high-quality mesh rendering on mobile devices. 

\subsection{Data Acquisition and Preprocessing}

Our objective is to develop a model capable of generating 3D renderings of an avatar.
Hence, in line with previous works ~\cite{saito2024rgca, ma2021pixel}, we utilize multi-view images captured from approximately 100 cameras.
Each subject is recorded performing a predefined set of facial expressions, gaze motions, and sentences.
Following the approach in ~\cite{saito2024rgca, ma2021pixel}, we conduct topologically consistent coarse mesh tracking and per-frame unwrapped averaged textures using the multi-view images.

\subsection{Encoder}

The encoder takes in the coarse mesh and the unwrapped averaged texture as input and produces a latent code $\mathbf{z}$. We use a network architecture similar to ~\cite{ma2021pixel}.
We only use this encoder during training since computing tracked meshes require high quality multi-view images.

\subsection{Decoder}

To make the avatar model animatable, we use a decoder $\mathcal{D}_m$  to produce mesh and texture, and a decoder $\mathcal{D}_g$ to produce  3D Gaussians. Both decoders are conditioned on the latent code. The decoders use a fully convolutional architecture similar to PiCA \cite{ma2021pixel}, for its efficiency to run on mobile devices. We produce both the mesh and Guassian parameters in the same UV space so that we can share compute and features across the convolutional layers in $\mathcal{D}_m$ and $\mathcal{D}_g$.

\subsection{Mesh Geometry}

The mesh provides the avatar's underlying geometry and visual information and guides the 3D Gaussians which are placed on it. We use a semi-transparent mesh to render the mesh with 3D Gaussians jointly. Although our method is independent of the underlying mesh model, we base our model on PiCA~\cite{ma2021pixel} for our experiments due to its ability to produce high-quality meshes and to run efficiently on mobile devices. 

The mesh decoder $\mathcal{D}_m$ takes the latent code $\mathbf{z}$ and viewing direction $\Theta_v$ as inputs, generating a texture and a geometry. All the outputs of the decoder are in the UV texture space producing a $K \times K $ map. The decoder outputs a color texture map $T_{c} \in R^{K \times K }$ in UV space, an opacity texture map $T_{\alpha} \in R^{K \times K }$, and a position map that contains vertices $X = \{ {x}_1, {x}_2 \dots {x}_{K^2} \}$ at each UV location, as follows:
\begin{equation}
\{ T_{c} ; T_{\alpha} ;  X \} = \mathcal{D}_m(\mathbf{z}; \Theta_v).
\end{equation}
The color texture decoder is conditioned on both the latent code and the viewing direction which helps the model in generating view-dependent appearance.
However, the geometry and transparency map are independent of the viewing angle, and hence, they are conditioned solely on the latent code.

The position map produced by the decoder along with a predefined human face topology $\tau$, is used to construct a mesh of the face.  This mesh is rendered using a differentiable renderer, allowing us to learn the whole system in an end-to-end manner. For a pixel $p$ in the rendered image, the mesh color $C'_p $, mesh opacity $ \alpha'_p $, and mesh depth $d'_p$ are produced by rasterizing the color texture $T_{c}$ and opacity texture $ T_{\alpha}$, 
\begin{align}
\{ C'_p, \alpha'_p, d'_p \} = \operatorname{rasterize}(X , \tau,    \{ T_{c}, T_{\alpha} \} , p  ).
\end{align}
Note that, in this stage, we render this mesh as an opaque surface with no blending, and therefore $\{ C'_p, \alpha'_p, d'_p \}$ correspond to the color, opacity, and depth of the front-most surface.
We also use edge gradients ~\cite{pidhorskyi2024rasterized} for improved optimization of the mesh.

\subsection{Gaussian Splatting}

\begin{figure}
\centering
\includegraphics [width=\linewidth]{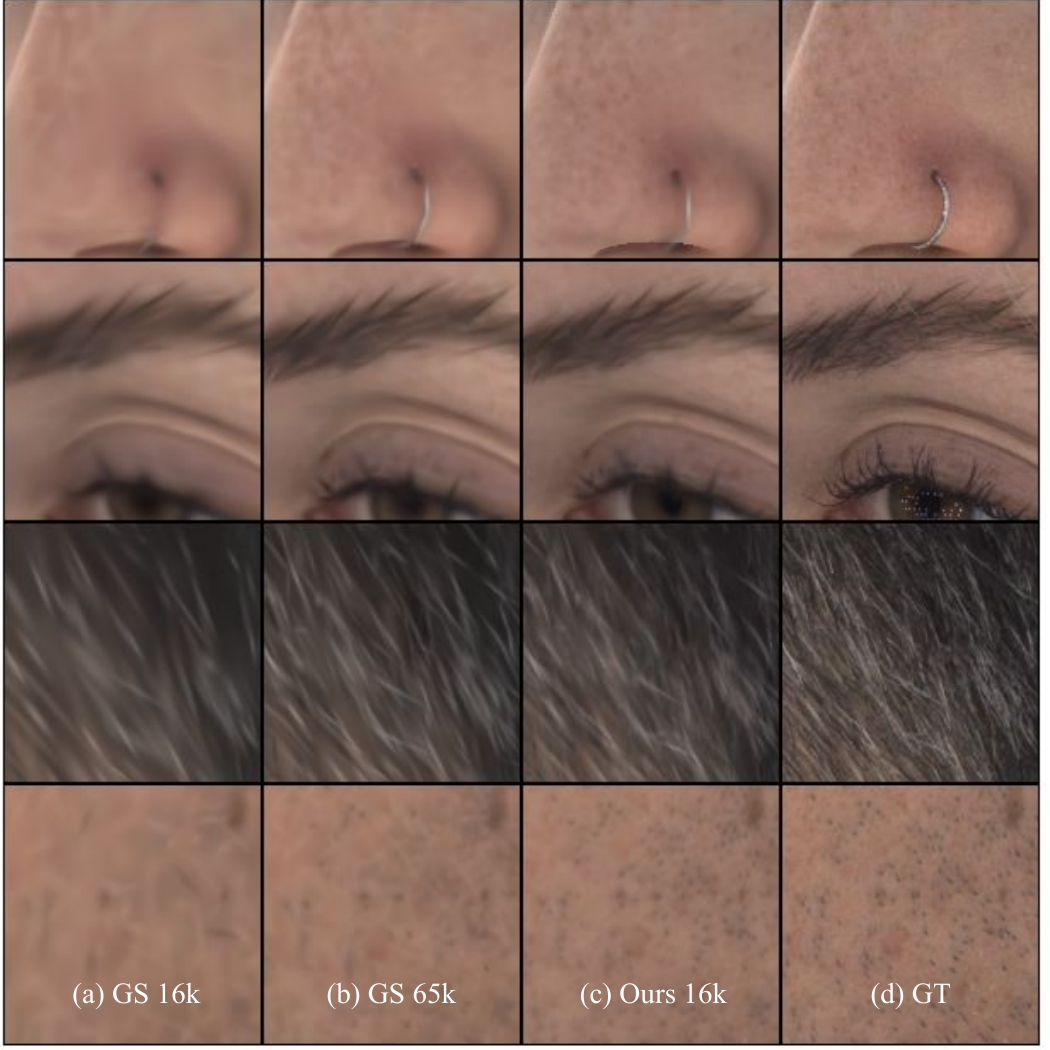}
\caption{\textbf{Comparing with vanilla Gaussian avatars.} With only 16k Gaussians, vanilla Gaussian avatars struggle to capture facial details. In contrast, our hybrid avatars, also using 16k 3D Gaussians, achieve significantly sharper representations. They are comparable to vanilla Gaussian avatars with 65k 3D Gaussians which are much slower to render. }  
\label{fig:compare_gs}
\end{figure}

\begin{figure}
\centering
\includegraphics[width=\linewidth]{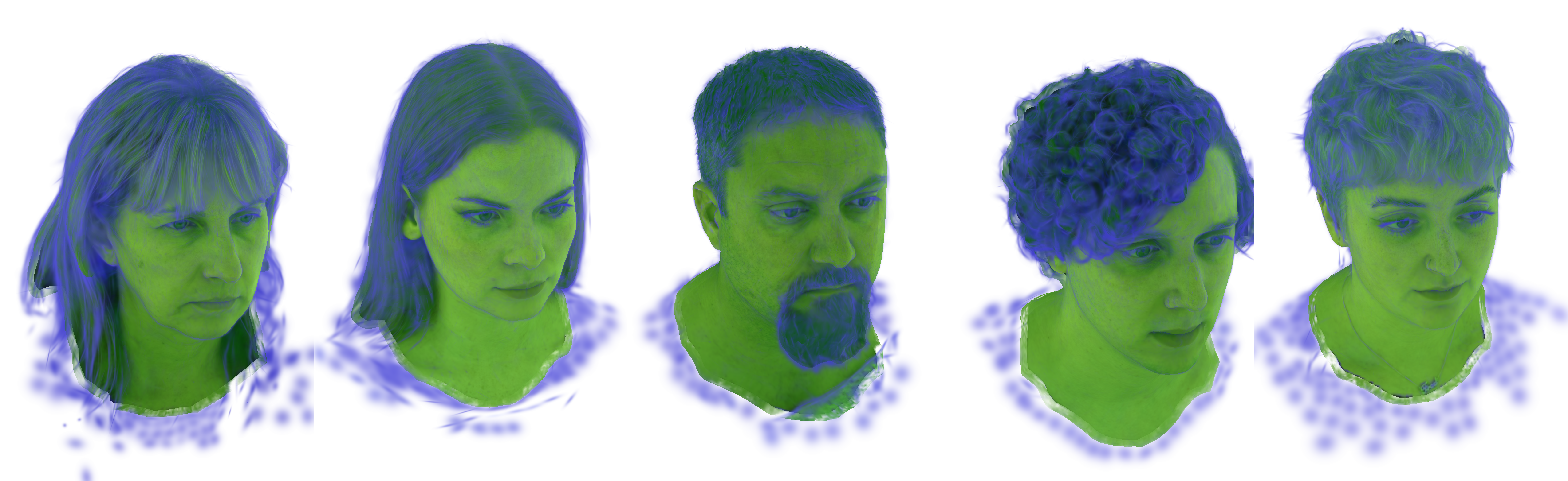}
\includegraphics[width=\linewidth]{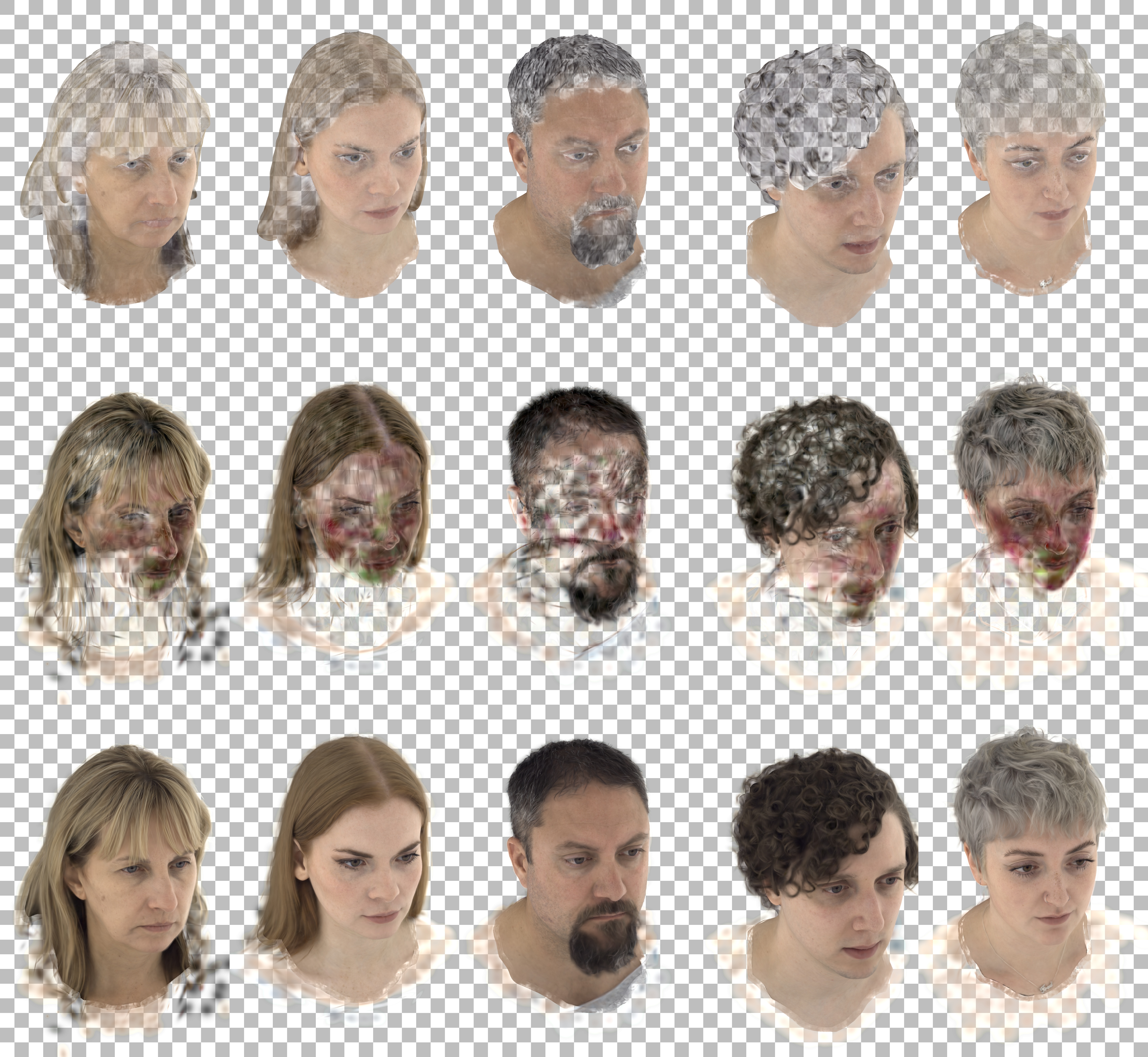}

\caption{\textbf{Mesh and Gaussian opacity contribution}. \textit{First row}: Mesh is colored green and Gaussians are colored  blue. \textit{Second row}: RGB contribution from the mesh. \textit{Third row}: RGB contribution from the Gaussians. \textit{Fourth row}: Final renders. 
}
\label{fig:splat_vs_mesh_assignment}
\end{figure}

\begin{figure}
\centering
\includegraphics [width=\linewidth]{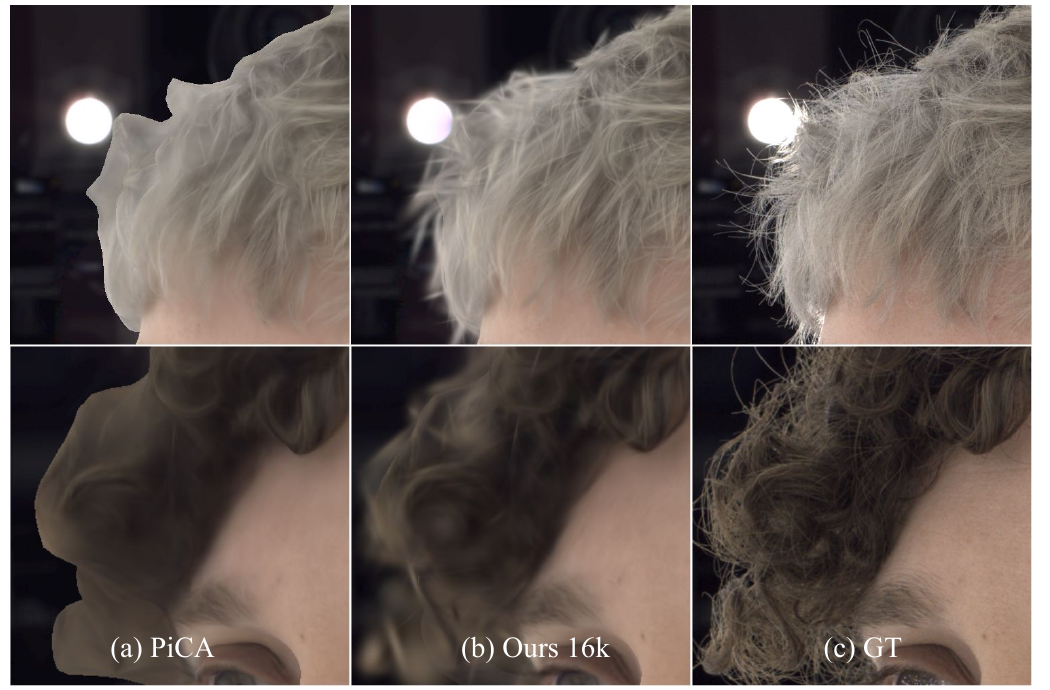}
\caption{\textbf{Comparing with mesh based avatars.} PiCA can only capture flat surfaces and struggles with complex areas like hair, where it coarsely approximates volumetric details. Our Hybrid GS avatars overcome this limitation by adding small number of 3D Gaussians. }  
\label{fig:compare_pica}
\end{figure}

Mesh-based models like PiCA~\cite{ma2021pixel} can create most facial features, but they struggle to produce high-quality hair, including head hair, facial hair, eyelashes, and eyebrows. On the other hand, 3D Gaussian Splatting is proficient at representing these elements. Hence, we use a set of 3D Gaussians along with the mesh to improve the quality in those regions.  For a set of $N$ 3D Gaussians, each individual Gaussian, denoted as $\mathbf{g}_k=\{\mathbf{t}_k, \mathbf{R}_k, \mathbf{s}_k, o, \mathbf{c}_k\}$, where $\mathbf{t}_k$ is the relative position, $\mathbf{R}_k \in SO(3)$ is the rotation, $\mathbf{s}_k$ is the scale, $o_k$ is the opacity, and $\mathbf{c}_k$ is the color.

We generate the 3D Gaussians by a decoder $\mathcal{D}_g$ that takes the latent code as input, and we share the convolutional neural network from the mesh decoder to produce the 3D Gaussian properties. Hence, the 3D Gaussians are produced in the same UV space, allowing feature re-use and shared computations.
\begin{equation}
\{ {g}_1, {g}_2 \dots {g}_N \} = \mathcal{D}_g(\mathbf{z}; \Theta_v)
\end{equation}
Since the mesh already captures the underlying geometry, we represent the positions of the 3D Gaussians relative to the mesh. The decoder generates the differences in positions with respect to the mesh coordinate positions.

To enhance rendering efficiency, we predict color directly from the decoder instead of using spherical harmonics. The predicted color is conditioned on the viewing angle and the latent code, helping in the production of view-dependent appearance.

We have limited budget on total number of 3D Gaussians that can be rendered efficiently on low-end devices. Hence, we want more 3D Gaussians in regions like hair, which cannot be represented by the mesh. For that, we use a static mask which selects 16,348 3D Gaussians from the $256 \times 256$ UV space. Out of the selected 3D Gaussians, 75\% of them are selected from the hair regions and the rest 25\% are selected from other regions. To compute the UV mask for the hair, we use semantic segmentation on the images and project them to the UV map.

\subsection{Hybrid Mesh-Gaussian Renderer}

After producing the mesh and the 3D Gaussians from the decoder, we jointly render them using our hybrid Mesh-Gaussian renderer. We use a two-pass approach where the mesh is first rendered separately, and then blend that with the Gaussians in 3D. 

For each 3D Gaussian $g_k$, let the depth for that gaussian in the camera space be $d_k$. We use the 3D Gaussians in the sorted order, and $d_1 \leq d_2 \dots \leq d_N $.

For every pixel $p$, we first accumulate all the 3D Gaussians which are in front of the mesh, then we accumulate the mesh color of that pixel, and finally accumulate the rest of the 3D Gaussians. 

If the depth of the mesh at pixel $p$ is $d'_p$, the color contribution of Gaussians in front of the mesh will be computed by accumulating $g_1, g_2 \dots g_{m-1} $, where $d_1 \leq d_2 \dots \leq d_{m-1} < d'_p $,
\begin{equation}
\label{eq:colorfront}
C_{front} =\sum_{k = 1}^{m-1} \mathbf{c}_k \alpha_k \prod_{j=1}^{k-1}\left(1-\alpha_j\right),
\end{equation} where the transparency $\alpha_k$ is evaluated using the 2D covariance of the Gaussian and multiplied by the per-Gaussian opacity $o_k$.

The color contribution of the mesh part will be computed using the color and opacity of the mesh at $p$ and the transmittance of the 3D Gaussians in front of it. 
\begin{equation}
\label{eq:colormesh}
C_{mesh} = {C'}_p \alpha'_p  \prod_{j=1}^{m-1}\left(1-\alpha_j\right) 
\end{equation}
The color contribution of Gaussians that are behind the mesh is computed by accumulating $g_m, g_{m+1} \dots g_N $, where $ d'_p  \leq d_m \leq d_{m+1} \dots \leq d_N,$
\begin{equation}
\label{eq:colorbehind}
C_{behind} = (1 - \alpha'_p) \sum_{k = m}^{N} \mathbf{c}_k \alpha_k \prod_{j=1}^{k-1}\left(1-\alpha_j\right).
\end{equation}
The final color at pixel $p$ is then 
\begin{equation}
\label{eq:colorfinal}
C_{p} = C_{front} + C_{mesh} + C_{behind}
\end{equation}
This makes it differentiable and lets us back propagate gradients to the mesh and Gaussians. By jointly training the decoder with the hybrid renderer, the network can decide what to represent using Gaussians and what to represent using mesh. Fig.~\ref{fig:splat_vs_mesh_assignment} shows how the mesh tends to explain flatter surfaces, like the skin, whereas the Gaussians tend to explain structures like hair and eyelashes.

\subsection{Optimization and regularization}

Given multi-view video data of a person, we jointly optimize all trainable network parameters. We use mean squared error loss between the rendered color $C_p$ and and ground truth color $C^{GT}_p$ on all the pixels of the image. The latent code $\mathbf{z}$ is computed from an encoder $\mathcal{E}$ . $\mathcal{E}$ takes the tracked mesh and an average texture computed by back-projecting ground truth images onto the tracked mesh as input.  $\mathcal{E}$ and the decoders $\mathcal{D}_g$ and $\mathcal{D}_m$ are trained jointly with a Kullback-Leibler divergence loss on the encoder outputs. 

We regularize the Gaussians with a scale loss~\cite{saito2024rgca} which penalizes if the scale term $s_k$ is not in a given range,
\begin{equation}
\begin{aligned}
\mathcal{L}_{\mathrm{s}} = \mathrm{mean}(l_{\mathrm{s}}), \, l_{\mathrm{s}} &= 
\begin{cases}
1/\max(s,10^{-7})  & \text{if } s < 0.1 \\
(s - 10.0)^2 & \text{if } s > 10.0\\
0 & \text{otherwise}.
\end{cases}
\end{aligned}
\end{equation}
We also ensure that the Gaussians are close to the mesh by using a loss to penalize if translation vector $t_k$ is very large,
\begin{equation}
\begin{aligned}
\mathcal{L}_{\mathrm{t}} = \mathrm{mean}(l_{\mathrm{t}}), \, l_{\mathrm{s}} &= 
\begin{cases}
\lVert t \lVert - 10.0 & \text{if } \lVert t \lVert > 10.0\\
0 & \text{otherwise}.
\end{cases}
\end{aligned}
\end{equation}
Additionally, to regularize the generated mesh, we use a normal loss and a Laplacian
smoothness regularization. Please refer to PiCA~\cite{ma2021pixel} for more details for the mesh regularization terms.

\begin{table}[]
\setlength{\tabcolsep}{4pt}
\begin{tabular}{|l|l|rrrr|}
\hline
\textbf{Subject}                 & \textbf{Method}        & \multicolumn{1}{l}{\textbf{MAE}} & \multicolumn{1}{l}{\textbf{SSIM}} & \multicolumn{1}{l}{\textbf{PSNR}} & \multicolumn{1}{l|}{\textbf{LPIPS}} \\ \hline
\multirow{4}{*}{Subject 1} & PiCA          & 9.13                    & 0.59                     & 26.05                    & 0.45                       \\
                        & GS 16k        & 8.15                    & 0.63                     & 27.24                    & 0.49                       \\
                        & GS 65k        & 7.67                    & \textbf{0.68}            & \textbf{27.85}           & 0.37                       \\
                        & Ours 16k & \textbf{7.65}           & 0.66                     & 27.68                    & \textbf{0.37}              \\ \hline
\multirow{4}{*}{Subject 2} & PiCA          & 6.48                    & 0.67                     & 28.76                    & 0.45                       \\
                        & GS 16k        & 6.37                    & 0.7                      & 29.3                     & 0.52                       \\
                        & GS 65k        & 5.8                     & \textbf{0.72}            & 30.02                    & 0.42                       \\
                        & Ours 16k & \textbf{5.7}            & 0.71                     & \textbf{30.12}           & \textbf{0.39}              \\ \hline
\multirow{4}{*}{Subject 3} & PiCA          & 6.59                    & 0.68                     & 28.18                    & 0.39                       \\
                        & GS 16k        & 6.65                    & 0.67                     & 28.53                    & 0.51                       \\
                        & GS 65k        & 6.29                    & 0.7                      & 28.96                    & 0.4                        \\
                        & Ours 16k & \textbf{6.13}           & \textbf{0.71}            & \textbf{29.06}           & \textbf{0.35}              \\ \hline
\multirow{4}{*}{Subject 4} & PiCA          & 6.72                    & 0.67                     & 28.13                    & 0.44                       \\
                        & GS 16k        & 6.53                    & 0.68                     & 28.77                    & 0.55                       \\
                        & GS 65k        & 6.79                    & 0.7                      & 28.77                    & 0.43                       \\
                        & Ours 16k & \textbf{6.04}           & \textbf{0.7}             & \textbf{29.25}           & \textbf{0.39}              \\ \hline
\multirow{4}{*}{Subject 5} & PiCA          & 7.33                    & 0.66                     & 27.54                    & 0.4                        \\
                        & GS 16k        & 6.88                    & 0.69                     & 28.37                    & 0.48                       \\
                        & GS 65k        & 6.48                    & 0.72                     & 28.95                    & 0.36                       \\
                        & Ours 16k & \textbf{6.21}           & \textbf{0.72}            & \textbf{29.18}           & \textbf{0.33}              \\ \hline
\end{tabular}
\caption{\textbf{Quantitative evaluation on face dataset.} Our proposed hybrid approach with 16k Gaussians outperforms PiCA (a pure mesh approach) and vanilla GS with 16k Gaussians, and is comparable to vanilla GS with 65k Gaussians.}
\label{table:main_quant}
\end{table}

\begin{table}[]
\begin{tabular}{ll|rrrr}
                                            &            & 8k & 16k & 32k & 65k \\ \hline
\multicolumn{1}{l|}{\multirow{2}{*}{LPIPS}} & Hybrid GS & 0.36     & 0.33      & 0.31      & 0.31      \\
\multicolumn{1}{l|}{}                       & Vanilla GS & 0.56     & 0.48      & 0.42      & 0.36      \\ \hline
\multicolumn{1}{l|}{\multirow{2}{*}{MAE}}   & Hybrid GS & 6.41     & 6.21      & 5.92      & 5.91      \\
\multicolumn{1}{l|}{}                       & Vanilla GS & 7.27     & 6.88      & 6.66      & 6.48      \\ \hline
\multicolumn{1}{l|}{\multirow{2}{*}{PSNR}}  & Hybrid GS & 28.82    & 29.18     & 29.56     & 29.57     \\
\multicolumn{1}{l|}{}                       & Vanilla GS & 27.92    & 28.37     & 28.67     & 28.95     \\ \hline
\end{tabular}
\caption{\textbf{Quality metrics with respect to number of Gaussians.} This table shows LPIPS, MAE, and PSNR metrics for hybrid and vanilla GS models for varying numbers of Gaussian splats.}
\label{table:numgaussiansmetrics}
\end{table}

\section{Experiments}

\subsection{Experimental Setting }

We evaluate our method on a dataset of human faces containing 5 subjects with diverse genders and hairstyles. We use 5 held-out camera poses that were not used during training to generate our qualitative and quantitative results. We used around 2,000 video frames of size 2048 $\times$ 1334 for training and evaluation. These frames contain several human expressions. We report MAE, PSNR, LPIPS~\cite{zhang2018perceptual}, and SSIM on the face region of the image. For a fair comparison, we train and evaluate all the compared models with the same quantity of data and the same number of iterations, and use the same base decoder architecture. 
Additionally, we also evaluate our method on a similar full body image dataset of a single subject.

\subsection{Baselines}
We compare our model with a mesh-based PiCA \cite{ma2021pixel} model, and purely 3D Gaussian Splatting based avatar models. The only difference between our hybrid model and the baseline Gaussian Splatting model is the absence of the mesh component in the rendering. Our GPiCA hybrid model uses 16,384 3D Gaussians, hence we compare against 16,384 and 65,536 (4$\times$) Gaussians vanilla 3D Gaussian Splatting based models. While even more Gaussians can be used to represent these avatars for better quality, we find that it gets prohibitively more expensive for low-end devices with integrated GPUs to render them at very high frame rates. We use the same network architecture for the decoders that produce the per-frame mesh and/or Gaussian primitives in all the models for a fair comparison. We also compare our method with opaque mesh hybrid Gaussian Splatting which is used by works such as ~\cite{svitov2024haha,wang2024mega}.

\subsection{Qualitative Results}

\begin{figure}
\centering
\includegraphics [width=\linewidth]{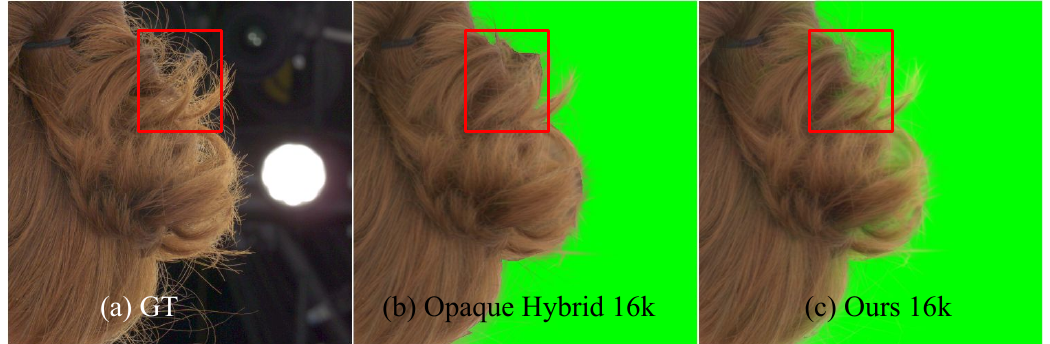}
\caption{\textbf{Comparing with opaque hybrid avatars.} Although opaque mesh hybrids show enhancements over pure mesh avatars, they still have limitations in complex regions like hair, where the mesh component coarsely approximates fine volumetric details. Our proposed hybrid avatars do not face these limitations. }  
\label{fig:opaque_vs_ours}
\end{figure}

Fig. \ref{fig:compare_gs} and \ref{fig:compare_pica} show renders of our model and the baselines from camera poses that are unseen during training. We can see that the mesh-only PiCA model fails to accurately represent hair and non-flat regions. The mesh-only model also produces sharp edges which makes it easily distinguishable from reality. Due to limited number of 3D Gaussians, the purely 3D Gaussian Splatting based model is not able to represent sharp details in face texture (skin spots, beard etc.), something which the PiCA model is better at, while being faster. On the other hand, our GPiCA hybrid model does not have the shortcomings of the baselines while using a limited number of 3D Gaussians.  GPiCA learns detailed skin textures, non-flat regions like hair, and also does not have any sharp edge artifacts from the mesh -- producing images closer to the ground truth. 
Other works, such as ~\cite{svitov2024haha,wang2024mega}, use an opaque mesh for blending with 3D Gaussians. 
Figure \ref{fig:opaque_vs_ours} shows that our proposed hybrid avatars perform better in regions like hair, which need complex volumetric modeling.
Hybrid 3D Gaussians with opaque mesh fail to model intricate volumetric details because coarse meshes tend to hinder 3D Gaussians that could go behind them, but our method does not have this limitation. 

\subsection{Quantitative Results}

Table \ref{table:main_quant} shows quantitative results of our model and its comparison with other baseline models. We can see that our hybrid model has better LPIPS and MAE scores compared to other baselines for all subjects, and better SSIM and PSNR scores for most subjects. Our hybrid model with 16,384 3D Gaussians is able to outperform vanilla 3DGS based models with same number of Gaussians and PiCA in all cases. Despite having 1/4th the number of 3D Gaussians, our model is able to get better LPIPS scores compared to the vanilla 3DGS based models with 65,536 3D Gaussians and performs competitively on other metrics. The results show that we can reduce the required number of Gaussians by 4$\times$ and effectively leverage mesh-based model for a lot of the heavy work, and still get similar or better quality outputs.

\subsection{Rendering Runtime}

We evaluate the mean GPU runtime performance of rendering PiCA, vanilla 3DGS models, and our hybrid GPiCA models in Table \ref{table:runtime}. We implemented efficient Vulkan based rendering modules for each of the models that run on a standard Quest 3 VR headset equipped with Adreno 740 GPU and a Hexagon Tensor Processor (HTP). The render resolution was kept to be 2048 $\times$ 1334 for all the numbers and Subject 5 models were used. We run the decoder on the HTP and the rendering on the GPU. The decoder for PiCA takes 2.2 ms, and the decoder for GS and hybrid models takes 6.9 ms. For PiCA, the rendering time includes for mesh rasterization and the multi-layer perceptron that runs in screen space in the fragment shader. For 3DGS, we implement a compute based rendering pipeline using Vulkan compute shaders. For GPiCA hybrid renderer, we modify the final color accumulation compute shader in the 3DGS pipeline to also take RGBA and depth buffers from the PiCA renderer as inputs. We then use these additional inputs for computing the final per-pixel color and opacity values from both the rasterized mesh and the Gaussians as described in Eq. \ref{eq:colorfinal}. Our hybrid renderer runtime is in the same range as the time for running PiCA renderer followed by 3DGS renderer, but the quality exceeds that of the 3DGS model with 4$\times$ Gaussians, the latter being much more expensive to render natively on VR devices with high refresh rates ($>$72 Hz).

\begin{table}
\centering
\begin{tabular}{|c|c|c|}
\hline
\textbf{Models} & \textbf{LPIPS}$\downarrow$ & \textbf{Render Time (ms)} \\
\hline
PiCA (Mesh) & 0.4 & 1.633\\
\hline
GS 16k & 0.48 & 9.0997\\
\hline
GS 65k & 0.36 & 19.266\\
\hline
Hybrid GS 16k & 0.33 & 10.900\\
\hline
\end{tabular}
\caption{\textbf{Performance and runtimes on Quest 3.} This table shows the LPIPS error metric and corresponding rendering times for various methods on a Quest 3 mobile GPU. Note that the Hybrid GS model with 16k splats has comparable or improved performance with a vanilla GS model of 65k splats with significantly improved runtime.}
\label{table:runtime}
\end{table}

\subsection{Further Analysis}

\subsubsection{Effect of number of Gaussians }
\begin{figure}
\centering
\includegraphics[width=\linewidth]{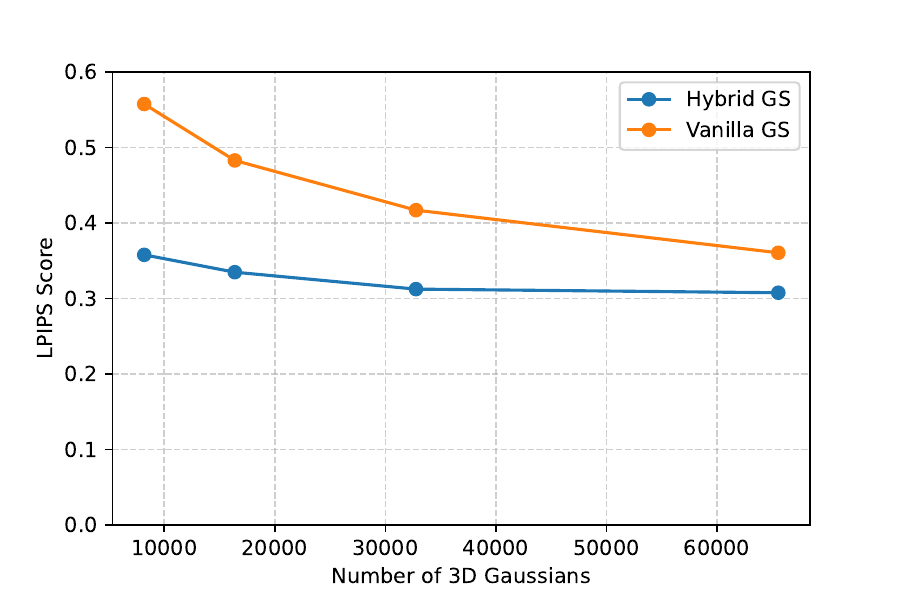}
\caption{\textbf{Number of Gaussians vs. LPIPS.} We show LPIPS reconstruction error for hybrid and vanilla GS models, for varying number of Gaussians. Note that a hybrid model with 8k splats performs as well as a vanilla GS model with 65K splats on this data.}
\label{fig:numgaussianslpips}
\end{figure}

Fig. \ref{fig:numgaussianslpips} and Table \ref{table:numgaussiansmetrics} show how the LPIPS scores of vanilla GS and our hybrid GS models change with the number of 3D Gaussians for Subject 5. 
We can see that LPIPS reduces when we increase the number of 3D Gaussians. Our hybrid GS based model is always better than the vanilla 3DGS base model for the same number of Gaussians, and better than the models with 2$\times$/4$\times$ Gaussians. We chose 16,384 Gaussians for our hybrid GS based model for most of the experiments as it is a good balance between runtime speed and quality. 

\subsubsection{Visualizing Geometry }
Fig. \ref{fig:normals} shows the visualizations of the geometry of the learned mesh and the 3D Gaussians for our hybrid GS model. We can see that the 3D Gaussians become long and elongated at the hair region learning finer hair details. The mesh geometry is smooth at the non-hair regions of the face and captures a coarse shape of the hair regions. 
Fig \ref{fig:splat_vs_mesh_assignment} shows a visualization of our hybrid GS model, highlighting the contributions of the mesh part (highlighted in green color) compared with the contributions from the Gaussians (highlighted in blue color). 
It also shows the learned mesh and Gaussians rendered independently. We can see how Gaussians complement the mesh at hair regions and composite together to generate the final colors.

\subsection{Ablations}

\textbf{Hybrid GS with uniform initialization.}
Rather than the proposed initialization where we put more 3D Gaussians in the hair region using segmentations, we initialize the locations of 3D Gaussians uniformly over the UV map. 

\noindent \textbf{Hybrid GS with opaque mesh.}
Rather than the proposed method where the opacity of the mesh surface layer is predicted by the decoder, we just set treat the mesh as fully opaque and set the opacity to 1.0. 

We show the performance metrics for both the above ablations in Table \ref{table:ablations} and observe a decrease in performance, which is expected. Opaque mesh would mean that the mesh colors are all accumulated before the Gaussians, which isn't necessarily true for avatars. With opaque mesh, it also fails to model intricate areas such as hair because the mesh coarsely approximates thin structures.  Uniform initialization of 3D Gaussians still performs well but makes it relatively more difficult to displace the Gaussians where they are needed the most.

\begin{table}[]
\setlength{\tabcolsep}{5pt}
\begin{tabular}{|l|c|c|c|c|}
        \hline
        \textbf{Model Name} & \textbf{MAE} & \textbf{SSIM} & \textbf{PSNR} & \textbf{LPIPS} \\ \hline
        GS 260K & 3.15 & 0.72 & 26.78 & 0.17 \\ \hline
        PiCA & \textbf{2.84} & \textbf{0.73} & 26.56 & 0.23 \\ \hline
        Opaque Hybrid 16k & 3.04 & 0.71 & 26.24 & 0.17 \\ \hline
        Ours 16k & 2.85 & \textbf{0.73} & \textbf{26.9} & \textbf{0.15} \\ \hline
    \end{tabular}
    
\caption{\textbf{Quantitative evaluation on full body dataset.} Our proposed hybrid approach with 16k 3D Gaussians outperforms vanilla Gaussian avatars while using 16x fewer 3D Gaussians. We also outperform PiCA and opaque mesh hybrid avatars.  }
\label{table:body_results}
\end{table}

\begin{table}[]
\setlength{\tabcolsep}{4pt}
\begin{tabular}{|l|rrrr|}
\hline
\textbf{Method}      & \multicolumn{1}{l}{\textbf{MAE}} & \multicolumn{1}{l}{\textbf{SSIM}} & \multicolumn{1}{l}{\textbf{PSNR}} & \multicolumn{1}{l|}{\textbf{LPIPS}} \\ \hline
Hybrid GS (Opaque) & 6.591                   & 0.688                    & 28.679                   & 0.374                      \\ \hline
Hybrid GS (Uniform)  & 6.432                   & 0.692                    & 28.951                   & 0.38                       \\ \hline
Hybrid GS      & \textbf{6.347}          & \textbf{0.699}           & \textbf{29.058}          & \textbf{0.367}             \\ \hline
\end{tabular}
\caption{\textbf{Ablations averaged across all 5 subjects.} In Hybrid GS Uniform we do not use a sampling mask to initialize  more 3D Gaussians at hair regions. In Hybrid GS Opaque, we use an opaque mesh rather than semi-transparent mesh. }
\label{table:ablations}
\end{table}

\section{Conclusion}

We present Gaussian Pixel Codec Avatars, a hybrid approach to decode and render avatars, leveraging the efficiency of mesh and the expressivity of 3D Gaussian Splatting in a unified pipeline. Unlike opaque mesh hybrid Gaussian Splatting approaches, our method is able to model thin and complex regions like hair, which is very important for avatars.
The hybrid renderer makes it possible to render very high quality avatars on low-compute, low-latency devices like VR headsets at their native refresh rates.

{
    \small
    \bibliographystyle{ieeenat_fullname}
    \bibliography{main}
}


\end{document}